\pdfoutput=1

\documentclass[11pt]{article}

\usepackage[final]{acl}

\usepackage{times}
\usepackage{latexsym}

\usepackage[T1]{fontenc}

\usepackage[utf8]{inputenc}

\usepackage{microtype}

\usepackage{inconsolata}

\usepackage{graphicx}

\usepackage{subcaption} 
\usepackage{booktabs} 

\title{How Well Do Large Language Models Disambiguate Swedish Words?}

\author{Richard Johansson \\
  Department of Computer Science and Engineering \\
  Chalmers University of Technology and University of Gothenburg \\
  \texttt{richard.johansson@gu.se}}

\begin{document}
\maketitle
\begin{abstract}
We evaluate a battery of recent large language models on two
benchmarks for word sense disambiguation in Swedish. At present, all
current models are 
less accurate than the best supervised disambiguators in 
cases where a training set is available, but most models outperform
graph-based unsupervised systems. Different prompting
approaches are compared, with a focus on how to express the set of possible
senses in a given context. The best accuracies are achieved when
human-written definitions of the senses are included in the prompts.
\end{abstract}

\section{Introduction}


Models of text and word meaning learned from corpus data have long
played an important role in computational lexical-semantic tasks such
as word sense disambiguation (WSD).
%
Nevertheless, while previous incarnations of language representation
models (e.g. count-based distributional vectors or learned static or
contextual word representations) have played a crucial role in many
approaches to WSD, there has been little work on
applying the latest 
generation of language models to this task. In particular, we are
aware of no previous work that investigates how well these models
perform for WSD in Swedish.



In this work, we evaluate several recent large language models (LLMs) on two
different evaluation sets. We consider the importance of what
information is provided in the prompts: in particular, we find that
the most effective prompts for most models use \emph{definitions} of
the senses, rather than just a set of related words, and we evaluate
different workarounds for situations where sense definitions are
unavailable.

While WSD has always been something of a niche topic within the wider
NLP field, it is probably useful to discuss why it is interesting at
all to consider this task at the current moment.
%
We see the importance of this study, and of WSD more generally
in the present day, as twofold: 1) rather than being an intermediate
step in an NLP pipeline, WSD is interesting \emph{in itself} in a
variety of use cases in 
lexical semantics; 2) it is also a useful for \emph{benchmarking} the
capabilities of modern language models, in particular for languages
other than English.








\section{The SALDO lexicon}
\label{sec:saldo}

The WSD experiments in this work are based on the SALDO
lexicon \cite{borin2013}, which defines a large sense inventory for
Swedish words. This lexicon is important in Swedish NLP since it is
used as a bridge between several lexical-semantic resources in
Swedish \cite{borin2010a}.

As discussed by \newcite{borin2009} and elsewhere, sense distinctions
in SALDO are comparatively coarse-grained.
Another fundamental difference 
to other well-known lexical-semantic resources is that SALDO does not
specify typed lexical-semantic relationships (such as synonymy or
hyponymy) between word senses but instead employs the concept of
association \cite{borin2013}, which can represent multiple types of
lexical-semantic relationships: often, an associated sense might be a
synonym or hypernym, but in other cases, it can be another relation
such as a meronym.

Although each sense could theoretically have association relationships
with many other senses, SALDO explicitly encodes connections between
each sense and its \emph{primary descriptor} (PD), an associated sense
with a more primitive meaning. Additional relationships are (more
irregularly) encoded as \emph{secondary descriptors}. Apart from these
relations, SALDO does not include any other lexical-semantic
information, such as sense definitions or contextual examples. Figure
1 illustrates the neighborhoods in the SALDO graph around two senses
of the noun \emph{ämne} (`substance' or `topic').

%
\begin{figure}[t]
  \centering
  \begin{subfigure}[b]{0.40\textwidth}
    \begin{center}
    \includegraphics[width=42mm]{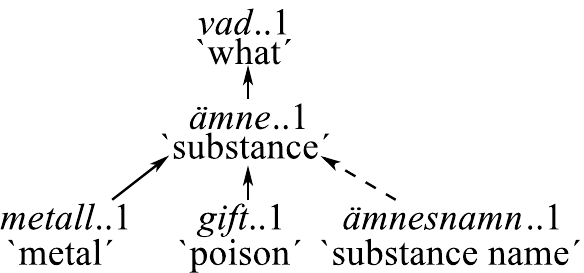}
    \caption{Neighborhood of \emph{ämne}..1 `substance'.}
    \label{subfig:amne1}
    \end{center}
  \end{subfigure}
  \begin{subfigure}[b]{0.40\textwidth}
    \begin{center}
    \includegraphics[width=42mm]{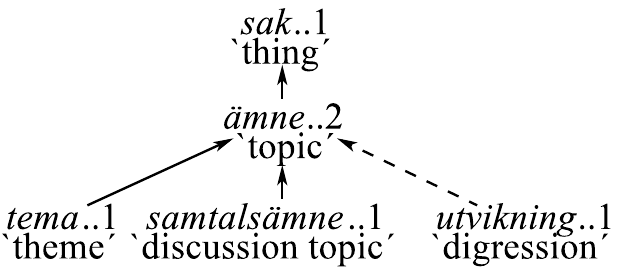}
    \caption{Neighborhood of \emph{ämne}..2 `topic'.}
    \end{center}
  \end{subfigure}

  \caption{Fragments of SALDO neighborhoods for two of the senses of
    \emph{ämne}. Primary descriptor edges are drawn as solid arrows
    and secondary descriptor edges as dashed arrows.
    \label{fig:saldoex}
  }
  
\end{figure}

\section{Method}

\subsection{Datasets and Experimental Setup}

The experiments are carried out on two different sense-annotated datasets.

\paragraph{SENSEVAL-2} The largest sense-annotated resource for
Swedish was developed in the SemTag project \cite{jarborg1999}; this
covers most of the SUC corpus \cite{ejerhed1992}.
The Swedish lexical sample of the \emph{SENSEVAL-2} shared
task \cite{kokkinakis2001} is a subset of the SUC dataset
and included annotated instances for 40 ambigous lemmas.
This dataset does not use the SALDO sense inventory, but 
the senses for these lemmas were manually mapped to SALDO by 
\newcite{nietopina2016a}.
Since SALDO uses a coarser division into
senses than SemTag, three of the lemmas were not ambiguous after the conversion and were removed from the dataset.

\paragraph{\emph{Eukalyptus}} The only running-text corpus annotated with SALDO senses is
\emph{Eukalyptus} \cite{johansson2016a}, which has texts from
eight domains.

\paragraph{Preprocessing} The instances were preprocessed using
the \emph{Sparv} pipeline \cite{borin2016}. For each word, the
pipeline proposes a set of possible SALDO senses, based on an
automatic morphological analysis and lemmatization.

For \textit{Eukalyptus}, 
unambiguous words were excluded from the experiment. This means that the \emph{practical} accuracy is higher than what we report in the next section, since the majority of the words are unambiguous.
We also exclude cases where the annotated sense is
a non-compositional reading of a multi-word expression (e.g. \emph{på
örat} intended as `drunk', not as `on the ear') or a compositional
reading of a compound.
After preprocessing the two datasets, SENSEVAL-2 consists of a test set of
1,366 instances and a training set of 7,790 instances,
and the \emph{Eukalyptus} set of 12,434 instances.

\paragraph{Experimental setup}
During evaluation, a disambiguator is given an instance that includes
the target word and a five-sentence context centered on the target
word. (Using larger contexts did not impact results meaningfully
except in terms of cost and processing time.)
The disambiguator then chooses one of the alternatives proposed by
the Sparv lemmatization pipeline.
%
%
For SENSEVAL-2, the reported accuracy is a macro-average over the 37
lemmas (20 nouns, 11 verbs, 6 adjectives). For \emph{Eukalyptus}, we
report a micro-averaged accuracy computed over all instances in the corpus.

\subsection{Baselines}

Before evaluating the LLM-based disambiguators, we ran a number of
trivial and nontrivial baselines on the same datasets. Tables~\ref{table:bl_senseval} and \ref{table:bl_eukalyptus} show the results on the SENSEVAL-2 and \emph{Eukalyptus} benchmarks, respectively.
The \emph{Random} accuracy corresponds to the mean accuracy achieved
when selecting a sense uniformly randomly from the set of
alternatives. The \emph{First sense} baseline selects the
the sense with the lowest numerical identifier among the alternatives;
while SALDO senses are not ordered by frequency, the most common
senses are often listed earlier, so this can be seen as a proxy of a
most-frequent-sense baseline. We also include \emph{Upper bound} here:
the accuracy we'd get if always selecting the gold-standard sense if it is
among the alternatives. Since there are occasional mistakes in
lemmatization and the list does not always include the true sense,
this is not exactly 1.0.

Among nontrivial baselines, we evaluated various \emph{graph-based unsupervised}
and \emph{supervised} methods.
Graph-based unsupervised methods use the SALDO graph in different ways
but are not trained on a sense-annotated dataset. In this category, we
evaluate three different systems: \emph{Personalized PageRank} uses a
graph algorithm to disambiguate \cite{agirre2009}; \emph{Sense vectors} is the method
by \newcite{johansson2015b}, which uses the SALDO graph to create
vector representations of senses. 
the \emph{BERT substitutes} method \cite{johansson2022} uses a
BERT model to propose words that could be substituted for the target
word, and then compares the substitute set to the graph neighborhoods
for the different senses. 

Among \emph{supervised} methods, we evaluated two approaches: a linear
SVM using a bag-of-words representation, and a logistic regression on a BERT
output at the target token. Both were implemented as ``word experts'' that use one
classifier per base form \cite{berleant1995}.
Supervised models were only evaluated for SENSEVAL-2, which comes with a train/test split; \emph{Eukalyptus} includes many low-frequency lemmas and a supervised word expert approach would be less meaningful.



\begin{table}[t]
\begin{center}
\begin{small}
\begin{tabular}{l c}
\toprule
\textbf{System} & \textbf{Accuracy} \\
\midrule
Random & 0.349 \\
First sense & 0.495 \\ 
Upper bound & 0.992 \\ \hline
Personalized PageRank & 0.497 \\ 
Sense vectors & 0.498 \\ 
BERT substitutes & 0.668 \\\hline
BERT + LR & 0.931 \\
BoW + SVM & 0.808 \\
\bottomrule
\end{tabular}
\end{small}
\end{center}
\caption{Macro-averaged accuracies on the SENSEVAL-2 test set for
three categories of baselines: trivial, graph-based unsupervised, and supervised models.}
\label{table:bl_senseval}
\end{table}

\begin{table}[t]
\begin{center}
\begin{small}
\begin{tabular}{l c}
\toprule
\textbf{System} & \textbf{Accuracy} \\
\midrule
Random & 0.402 \\
First sense & 0.658 \\ 
Upper bound & 0.992 \\ \hline
BERT-based substitutes & 0.702 \\
\bottomrule
\end{tabular}
\end{small}
\end{center}
\caption{Micro-averaged accuracies on the \emph{Eukalyptus}
test set for two categories of baselines: trivial and graph-based unsupervised models.}
\label{table:bl_eukalyptus}
\end{table}

\subsection{Included Models}
In our experiments, we considered the following models. They were accessed through their respective APIs without any fine-tuning.

\begin{itemize}\addtolength{\itemsep}{-0.5\baselineskip}
\item \emph{Claude 3.5 Sonnet} \cite{anthropic2024};
\item \emph{Command R+} (170B) \cite{cohere2024};
\item \emph{Gemini 1.5 Pro} \cite{geminiteam2024};
\item \emph{Llama 3 8B} and \emph{70B}, and \emph{Llama 3.1 405B} \cite{llama2024}, via the Replicate API;\footnote{\url{https://replicate.com/home}}
\item \emph{GPT-3.5 Turbo} \cite{ouyang2022}, \emph{GPT-4
Turbo} \cite{openai2023}, \emph{GPT-4o}, and \emph{GPT-4o-mini}.
\end{itemize}
Appendix~\ref{app:apicosts} exemplifies usage costs for some of these models. 

\subsection{Prompt Design}
The prompts consisted of the following parts:
\begin{enumerate}\addtolength{\itemsep}{-0.65\baselineskip}
\item a preamble describing the WSD task: that the goal is to pick one out of a given list of senses;
\item an example demonstration;
\item a list of senses to choose from;
\item the context to disambiguate.
\end{enumerate}
The investigations in the next section focus on the
third point: how the senses are specified.

\section{Experiments}

\subsection{Neighborhoods or Definitions?}
\label{ss:bigeval}

In our first investigation, we investigated different ways of
presenting the list of SALDO senses to the model.
As mentioned in \S\ref{sec:saldo}, the only information about senses given directly in SALDO is the set of neighbors in the sense graph, as in Figure~\ref{fig:saldoex}.

The first prompting approach, the \emph{neighborhood} prompting technique, specifies the sense inventory for a given lemma simply by enumerating the lemmas of up to four direct neighbors in the sense graph. For a sense $s$, we first list the primary descriptor of $s$, followed by senses for which $s$ is the primary descriptor. We also include secondary descriptors if the number of neighbors is less than 4.
For instance, the neighborhood of the first sense of \emph{ämne} `substance' shown in Figure~\ref{subfig:amne1} would be encoded as \emph{vad, metall, gift, ämnesnamn}.
Appendix~\ref{app:nbprompt} shows an example in detail.

We compare the neighborhood prompting technique to a second approach where we give textual definitions of the senses.
Definitions are not available in SALDO, but for the SENSEVAL-2 dataset we can rely on the mapping between SALDO and SemTag senses and use their definitions.
Appendix~\ref{app:defprompt} shows how these prompts are written.

Table~\ref{table:senseval} shows the results on the SENSEVAL-2 test set for all models with the two prompting techniques. 
To give an impression of the variation in accuracy across the 37
lemmas, detailed results for the best LLM, a trivial baseline, and the best supervised model are presented in Appendix~\ref{app:detailed}.
The takeaways from this evaluation are: 1) \emph{all} 10 LLMs perform better with definition prompts than with neighborhoods, although there is some varition in the performance gap between the two approaches; 2) unsurprisingly, newer models outperform older models and larger models perform smaller models; 3) most LLMs outperform the graph-based unsupervised baselines; 4) no LLM reaches the level of the best supervised baseline.

\begin{table}[t]
\begin{center}
\begin{small}
\begin{tabular}{l cc}
\toprule
\textbf{Model} & \textbf{N} & \textbf{D} \\
\midrule
\texttt{claude-3-5-sonnet} & 0.778 & 0.855 \\
\texttt{gpt-4o} & 0.792 & 0.849 \\
\texttt{llama-3.1-405b-instruct} & 0.728 & 0.818 \\
\texttt{gpt-4-turbo} & 0.745 & 0.805 \\
\texttt{llama-3-70b-instruct} & 0.699 & 0.767 \\
\texttt{gemini-1.5-pro} & 0.730 & 0.737 \\
\texttt{gpt-4o-mini} & 0.676  & 0.727 \\
\texttt{gpt-3.5-turbo} & 0.466 & 0.551 \\
\texttt{command-r-plus} & 0.431 & 0.527 \\
\texttt{llama-3-8b-instruct} & 0.363 & 0.522 \\
\bottomrule
\end{tabular}
\end{small}
\end{center}
\caption{SENSEVAL-2 accuracies for all models. We compare prompts based on SALDO neighborhoods (N) and definitions (D).}
\label{table:senseval}
\end{table}

\begin{table}[t]
\begin{center}
\begin{small}
\begin{tabular}{l ccc}
\toprule
\textbf{Model} & \textbf{N} & \textbf{AD} & \textbf{CoT} \\
\midrule
\texttt{gpt-4o} & 0.792 & 0.792 & 0.788 \\ 
\texttt{llama-3.1-405b-instruct} & 0.728 & 0.788 & 0.758 \\
\texttt{llama-3-70b-instruct} &  0.699 & 0.767 & 0.737 \\
\texttt{gpt-4o-mini} & 0.676 & 0.695 & 0.739 \\
\bottomrule
\end{tabular}
\end{small}
\end{center}
\caption{SENSEVAL-2 accuracies with neighborhoods (N), automatically generated definitions (AD), and chain-of-thought prompts (CoT).}
\label{table:adresults}
\end{table}

\subsection{Model-written Sense Definitions}

The results in \S\ref{ss:bigeval} showed that all LLMs perform better with explicit definitions of senses rather than implicit specifications via SALDO neighborhoods.
However, since the SALDO--SemTag mapping is only definied for the 37 lemmas in the SENSEVAL-2 benchmark, we are unable to use definition-based prompts in the wild. But could we work with definitions written \emph{automatically} by a model instead of lexicographer-written definitions?

We implemented this idea in two different ways. First, we let GPT-4o write automatically generated sense definitions (AD) based on SALDO neighborhoods and used them as in the previous definition-based prompts. (Appendix~\ref{app:writedefprompt} shows the prompt we used to write the definitions.) We did not evaluate definitions written by any other model.
Secondly, we used a \emph{chain-of-thought} (CoT) approach \cite{wei2022} where the model is given the SALDO neighborhoods of the senses for a given instance, and is asked to write definitions before selecting the sense identifier.
For reasons of cost and computational efficiency, we only considered a subset of models in this investigation.

Table~\ref{table:adresults} shows the SENSEVAL-2 results. The best-perfoming model (GPT-4o) is not improved by any of these techniques: the scores are more or less unchanged from the baseline (neighborhood prompts). For the weaker models, both variants of definition-writing approaches seem to improve over the baseline. There is no consistent difference over the set of models between AD and CoT.

In addition, we compared neighborhoods and GPT-4o-written definitions
on the larger \emph{Eukalyptus} dataset, and
Table~\ref{table:eukresults} shows the accuracies. CoT is not
evaluated here because of its higher cost: with AD, each sense only
has to be defined once. Compared to SENSEVAL-2, \emph{Eukalyptus} has
a different distribution of lemmas and senses, and the results are
different from those in Table~\ref{table:adresults}. While the
relative ranking of models is the same, it does not seem to be useful
to include automatically written definitions in the
prompts (except for Llama-70B).

\begin{table}[t]
\begin{center}
\begin{small}
\begin{tabular}{l cc}
\toprule
\textbf{Model} & \textbf{N} & \textbf{AD} \\
\midrule
\texttt{gpt-4o} & 0.852 & 0.818 \\
\texttt{llama-3.1-405b-instruct} & 0.826 & 0.823 \\
\texttt{llama-3-70b-instruct} & 0.781 & 0.799 \\
\texttt{gpt-4o-mini} & 0.756  & 0.756 \\
\bottomrule
\end{tabular}
\end{small}
\end{center}
\caption{\textit{Eukalyptus} accuracies with neighborhoods (N) and automatically generated definitions (AD).}
\label{table:eukresults}
\end{table}


\section{Conclusions}


To answer the question posed by the title of this paper,
current LLM word sense disambiguators without any fine-tuning
perform at a level between unsupervised graph-based and
supervised disambiguators for the Swedish datasets we considered. All
LLMs perform better in this task when textual \emph{definitions} are
used to specify the senses, as opposed to just listing a set of
related words; however, even with the neighborhood-based approach, 
disambiguation accuracy is still substantially better than with
unsupervised approaches for most LLMs. Definitions written
automatically seem less useful than human-written definitions and do
not consistently outperform neighborhood-based prompts.

Based on these results, we can conclude that most recent LLMs have
some capability of carrying out a task that requires a fairly
fine-grained semantic representation of passages in Swedish. It would
be useful to consider the amount of Swedish included in pre-training and
instruction-tuning for these models, and how these quantities relate
to the performance figures reported here, but as of now this
information is unavailable.

\bibliography{custom}

\begin{thebibliography}{20}
\providecommand{\natexlab}[1]{#1}

\bibitem[{Agirre and Soroa(2009)}]{agirre2009}
Eneko Agirre and Aitor Soroa. 2009.
\newblock \href {https://aclanthology.org/E09-1005} {Personalizing {P}age{R}ank
  for word sense disambiguation}.
\newblock In \emph{Proceedings of the 12th Conference of the {E}uropean Chapter
  of the {ACL} ({EACL} 2009)}, pages 33--41, Athens, Greece.

\bibitem[{Anthropic(2024)}]{anthropic2024}
Anthropic. 2024.
\newblock \href
  {https://www-cdn.anthropic.com/fed9cc193a14b84131812372d8d5857f8f304c52/Model_Card_Claude_3_Addendum.pdf}
  {Claude 3.5 {S}onnet model card addendum}.

\bibitem[{Berleant(1995)}]{berleant1995}
Daniel Berleant. 1995.
\newblock \href
  {https://www.cambridge.org/core/journals/natural-language-engineering/article/engineering-word-experts-for-word-disambiguation/2F7A3A4E468C758870CED88DCD38A5AF}
  {Engineering ``word experts'' for word disambiguation}.
\newblock \emph{Natural Language Engineering}, 1:339--362.

\bibitem[{Borin et~al.(2010)Borin, Dann\'ells, Forsberg, Gronostaj, and
  Kokkinakis}]{borin2010a}
Lars Borin, Dana Dann\'ells, Markus Forsberg, Maria~Toporowska Gronostaj, and
  Dimitrios Kokkinakis. 2010.
\newblock \href
  {https://euralex.org/publications/the-past-meets-the-present-in-swedish-framenet/}
  {The past meets the present in the {S}wedish {F}rame{N}et++}.
\newblock In \emph{Proceedings of EURALEX}.

\bibitem[{Borin and Forsberg(2009)}]{borin2009}
Lars Borin and Markus Forsberg. 2009.
\newblock \href {http://hdl.handle.net/10062/9836} {All in the family: A
  comparison of {SALDO} and {WordNet}}.
\newblock In \emph{Proceedings of the Nodalida 2009 Workshop on WordNets and
  other Lexical Semantic Resources - between Lexical Semantics, Lexicography,
  Terminology and Formal Ontologies. NEALT Proceedings Series}, volume~7.

\bibitem[{Borin et~al.(2016)Borin, Forsberg, Hammarstedt, Ros\'{e}n, Schäfer,
  and Schumacher}]{borin2016}
Lars Borin, Markus Forsberg, Martin Hammarstedt, Dan Ros\'{e}n, Roland
  Schäfer, and Anne Schumacher. 2016.
\newblock \href
  {https://people.cs.umu.se/johanna/sltc2016/abstracts/SLTC_2016_paper_31.pdf}
  {Sparv: {S}pråkbanken's corpus annotation pipeline infrastructure}.
\newblock In \emph{Swedish Language Technology Conference}, Umeå, Sweden.

\bibitem[{Borin et~al.(2013)Borin, Forsberg, and L\"onngren}]{borin2013}
Lars Borin, Markus Forsberg, and Lennart L\"onngren. 2013.
\newblock \href {https://doi.org/10.1007/s10579-013-9233-4} {{SALDO}: a touch
  of yin to {WordNet}'s yang}.
\newblock \emph{Language Resources and Evaluation}, 47(4):1191--1211.

\bibitem[{Cohere(2024)}]{cohere2024}
Cohere. 2024.
\newblock \href {https://docs.cohere.com/docs/command-r-plus#model-details}
  {Command {R+}}.

\bibitem[{Ejerhed et~al.(1992)Ejerhed, K{\"a}llgren, Wennstedt, and
  {\AA}str{\"o}m}]{ejerhed1992}
Eva Ejerhed, Gunnel K{\"a}llgren, Ola Wennstedt, and Magnus {\AA}str{\"o}m.
  1992.
\newblock \href {https://dl.acm.org/doi/10.3115/976744.976808} {The linguistic
  annotation system of the {Stockholm-Ume{\aa}} corpus project -- description
  and guidelines}.
\newblock Technical report, Department of Linguistics, Ume\aa{} University.

\bibitem[{Google(2024)}]{geminiteam2024}
{Gemini Team,} Google. 2024.
\newblock \href {https://arxiv.org/pdf/2403.05530} {Gemini 1.5: Unlocking
  multimodal understanding across millions of tokens of context}.
\newblock \emph{arXiv preprint arXiv:2403.05530}.

\bibitem[{Johansson(2022)}]{johansson2022}
Richard Johansson. 2022.
\newblock \href {https://gupea.ub.gu.se/handle/2077/74254} {Coveting your
  neighbor's wife: Using lexical neighborhoods in substitution-based word sense
  disambiguation}.
\newblock In Elena Volodina, Dana Dann\'{e}lls, Aleksandrs Berdicevskis, Markus
  Forsberg, and Shafqat Virk, editors, \emph{LIVE and LEARN -- Festschrift in
  honor of Lars Borin}, pages 61--66. University of Gothenburg, Gothenburg,
  Sweden.

\bibitem[{Johansson et~al.(2016)Johansson, Adesam, Bouma, and
  Hedberg}]{johansson2016a}
Richard Johansson, Yvonne Adesam, Gerlof Bouma, and Karin Hedberg. 2016.
\newblock \href {https://aclanthology.org/L16-1482/} {A multi-domain corpus of
  {S}wedish word sense annotation}.
\newblock In \emph{Proceedings of the Language Resources and Evaluation
  Conference (LREC)}, pages 3019--3022, Portoro\v{z}, Slovenia.

\bibitem[{Johansson and {Nieto Pi\~{n}a}(2015)}]{johansson2015b}
Richard Johansson and Luis {Nieto Pi\~{n}a}. 2015.
\newblock \href {http://www.aclweb.org/anthology/W/W15/W15-1811.pdf} {Combining
  relational and distributional knowledge for word sense disambiguation}.
\newblock In \emph{Proceedings of the 20th Nordic Conference of Computational
  Linguistics}, pages 69--78, Vilnius, Lithuania. Link\"oping University
  Electronic Press, Sweden.

\bibitem[{Järborg(1999)}]{jarborg1999}
Jerker Järborg. 1999.
\newblock \href
  {http://spraakdata.gu.se/publikationer/guiss/GU-ISS-1999-06.pdf} {Lexikon i
  konfrontation}.
\newblock Technical report, University of Gothenburg.
\newblock Research Reports from the Department of Swedish, Språkdata,
  GU-ISS-99-6.

\bibitem[{Kokkinakis et~al.(2001)Kokkinakis, J\"{a}rborg, and
  Cederholm}]{kokkinakis2001}
Dimitrios Kokkinakis, Jerker J\"{a}rborg, and Yvonne Cederholm. 2001.
\newblock \href {https://aclanthology.org/S01-1011/} {{SENSEVAL}-2: The
  {S}wedish framework}.
\newblock In \emph{Proceedings of SENSEVAL-2 Second International Workshop on
  Evaluating Word Sense Disambiguation Systems}, pages 45--48, Toulouse,
  France.

\bibitem[{Llama~Team(2024)}]{llama2024}
AI~@~Meta Llama~Team. 2024.
\newblock \href {https://arxiv.org/pdf/2407.21783} {The {L}lama 3 herd of
  models}.
\newblock \emph{arXiv preprint arXiv:2407.21783}.

\bibitem[{{Nieto Pi\~{n}a} and Johansson(2016)}]{nietopina2016a}
Luis {Nieto Pi\~{n}a} and Richard Johansson. 2016.
\newblock Benchmarking word sense disambiguation systems for {S}wedish.
\newblock In \emph{Swedish Language Technology Conference}, Umeå, Sweden.

\bibitem[{OpenAI(2023)}]{openai2023}
OpenAI. 2023.
\newblock \href {https://arxiv.org/abs/2303.08774} {{GPT}-4 technical report}.
\newblock \emph{arXiv preprint arXiv:2303.08774}.

\bibitem[{Ouyang et~al.(2022)Ouyang, Wu, and {Jiang et al.}}]{ouyang2022}
Long Ouyang, Jeff Wu, and Xu~{Jiang et al.} 2022.
\newblock \href {https://arxiv.org/abs/2203.02155} {Training language models to
  follow instructions with human feedback}.
\newblock \emph{arXiv preprint arXiv:2203.02155}.

\bibitem[{Wei et~al.(2022)Wei, Wang, Schuurmans, Bosma, Ichter, Xia, Chi, Le,
  and Zhou}]{wei2022}
Jason Wei, Xuezhi Wang, Dale Schuurmans, Maarten Bosma, Brian Ichter, Fei Xia,
  Ed~H. Chi, Quoc~V. Le, and Denny Zhou. 2022.
\newblock \href {https://dl.acm.org/doi/10.5555/3600270.3602070}
  {Chain-of-thought prompting elicits reasoning in large language models}.
\newblock In \emph{Proceedings of the 36th International Conference on Neural
  Information Processing Systems}, NIPS '22, Red Hook, NY, USA. Curran
  Associates Inc.

\end{thebibliography}

\newpage

\appendix
\onecolumn

\section{Example of Detailed SENSEVAL-2 Results}
\label{app:detailed}

Table \ref{table:detailed} shows the evaluation scores for each lemma in the SENSEVAL-2 test set of the best-performing LLM (Claude 3.5 Sonnet) with definition-based prompts, the first-sense baseline, and the best-performing supervised system (BERT with logistic regression).
\begin{table}[h]
\begin{tabular}{lcccc}
\toprule
\textbf{Lemma} & $N$ & \textbf{Claude} & \textbf{Baseline} & \textbf{BERT+LR} \\
\midrule
\emph{barn}&                115 & 0.8348 & 0.5826 & 0.8783 \\
\emph{betydelse}&            52 & 1.0000 & 0.1538 & 0.9808 \\
\emph{bred}&                 18 & 1.0000 & 0.3889 & 1.0000 \\
\emph{flytta}&               32 & 0.8438 & 0.3125 & 0.9375 \\
\emph{fylla}&                11 & 1.0000 & 1.0000 & 1.0000 \\
\emph{färg}&                 19 & 0.9474 & 0.6316 & 0.8947 \\
\emph{förklara}&             30 & 0.9000 & 0.5667 & 0.9000 \\
\emph{gälla}&               148 & 0.7635 & 0.6216 & 0.9459 \\
\emph{handla}&               44 & 1.0000 & 0.1136 & 1.0000 \\
\emph{höra}&                 92 & 0.8696 & 0.3370 & 0.8696 \\
\emph{klar}&                 54 & 0.9259 & 0.1296 & 0.9630 \\
\emph{konst}&                13 & 0.6923 & 0.4615 & 0.7692 \\
\emph{kraft}&                20 & 0.9000 & 0.8500 & 1.0000 \\
\emph{kyrka}&                27 & 0.9259 & 0.7037 & 0.9630 \\
\emph{känsla}&               25 & 0.6400 & 0.6800 & 0.9600 \\
\emph{ledning}&              16 & 0.4375 & 0.5625 & 1.0000 \\
\emph{makt}&                 21 & 1.0000 & 1.0000 & 1.0000 \\
\emph{massa}&                16 & 0.9375 & 0.5625 & 0.9375 \\
\emph{mening}&               28 & 1.0000 & 0.5357 & 0.9643 \\
\emph{måla}&                 16 & 0.8125 & 0.2500 & 0.8750 \\
\emph{natur}&                16 & 1.0000 & 0.1250 & 0.9375 \\
\emph{naturlig}&             24 & 0.8750 & 0.3750 & 0.8750 \\
\emph{program}&              24 & 0.5417 & 0.5000 & 1.0000 \\
\emph{rad}&                  25 & 0.8000 & 0.2400 & 0.9200 \\
\emph{rum}&                  39 & 0.9487 & 0.8974 & 1.0000 \\
\emph{scen}&                 17 & 0.8235 & 0.7059 & 0.9412 \\
\emph{skjuta}&               12 & 0.7500 & 0.3333 & 0.8333 \\
\emph{spela}&                38 & 0.8947 & 0.2895 & 0.9211 \\
\emph{stark}&                62 & 0.5968 & 0.4194 & 0.8387 \\
\emph{tillfälle}&            20 & 0.9500 & 0.6000 & 0.9500 \\
\emph{uppgift}&              30 & 1.0000 & 0.1667 & 0.9667 \\
\emph{vatten}&               50 & 0.8800 & 0.8000 & 0.9800 \\
\emph{vänta}&                43 & 0.7907 & 0.6512 & 0.9070 \\
\emph{ämne}&                 34 & 0.9118 & 0.1765 & 0.8529 \\
\emph{öka}&                  77 & 0.7792 & 0.6364 & 0.9610 \\
\emph{öppen}&                33 & 0.8485 & 0.2424 & 0.7576 \\
\emph{öppna}&                25 & 0.8000 & 0.7200 & 0.9600 \\
\bottomrule
\end{tabular}
\caption{Detailed evaluation scores on the SENSEVAL-2 test set.}
\label{table:detailed}
\end{table}

\section{Prompts}
\label{app:prompts}

\subsection{Neighborhood-based prompt}
\label{app:nbprompt}
\begin{verbatim}
SYSTEM PROMPT:
You are a tool that carries out word sense disambiguation in Swedish. In the following, you will
be given sentences where your task is to disambiguate the word surrounded by asterisks (*). You
should select a sense identifier from a given list of senses. Each sense is specified by a set of
related words. Based on the related words, first write definitions of the senses, then explain
how the example relates to one of the senses, and finally output the relevant sense of the word
in the context.

Answer with one of the sense identifiers, or 0 if none of them are applicable. The last line of
the output must correspond to the sense identifier and include no other text.

Example input:
Entry: rock
Senses:  
1: related to "kappa", "bilrock", “bonjour”; 2: related to "musik", "hårdrock", "indierock".
Sentence: Bandet spelade * rock * .
Output:
2

USER PROMPT:
Entry: öppna
Senses: 1: related to "öppen", "bryta", "bryta upp", "dekantera";
2: related to "starta", "öppnande", "verksamhet"
Sentence: " Ja , du får nog göra det " , sa en av dem . Söderberg såg oförstående på honom . " Du
får nog ta och * öppna * " , sa mannen igen . Söderberg ville krympa ihop och bli liten . Så liten
att han försvann .
\end{verbatim}

\subsection{Definition-based prompt}
\label{app:defprompt}
\begin{verbatim}
SYSTEM PROMPT:
You are a tool that carries out word sense disambiguation in Swedish. In the following, you will
be given sentences where your task is to disambiguate the word surrounded by asterisks (*). You
should select a sense identifier from a given list of senses. Based on the definitions of the
senses, first explain how the example relates to one of the senses, and finally output the 
relevant sense of the word in the context.
Answer with one of the sense identifiers, or 0 if none of them are applicable. The last line of 
the output must correspond to the sense identifier and include no other text.

Example input:
Entry: rock
Sense definitions:
1: ytterplagg för överkroppen, som räcker ungefär till knäet och har ärmar
2: typ av melodisk enkel musik i kraftfullt markerad 4/4-takt
Sentence: Bandet spelade * rock * .
Output:
2

USER PROMPT:
Entry: öppna
Sense definitions:
1: bringa till öppet (eller öppnare) läge; med avs. på spärrande anordning, ibl. underförstådd; 
äv. med tanke på utrymmet innanför, personen etc. som skall släppas in m.m.; ibl. symboliskt i 
liknelser
2: göra tillgänglig för användande eller utnyttjande; av större grupp personer; med avs. på 
inrättning e. d., ibl. underförstådd; äv. med avs. på tidpunkt; inrätta (och påbörja) verksamhet 
med; ngt slags företag e. d.; börja utföra; viss angiven el. underförstådd verksamhet
Sentence: " Ja , du får nog göra det " , sa en av dem . Söderberg såg oförstående på honom . " Du 
får nog ta och * öppna * " , sa mannen igen . Söderberg ville krympa ihop och bli liten . Så liten
att han försvann .
\end{verbatim}

\subsection{Prompt for Writing Definitions}
\label{app:writedefprompt}

\begin{verbatim}
SYSTEM PROMPT:
You are lexicographer who writes dictionaries describing words in Swedish.

In the following, you will be given a list of senses of a given input word. For each sense, a set
of related words is given. Your task is to write a short definition in Swedish and give an example
sentence of each of the given senses.

The output should be a JSON object where each key is a sense identifier and each value a list
containing the definition and the example sentence. 

Example input:
Entry: "rock"
Senses:  1: related to "kappa", "bilrock", “bonjour”; 2: related to "musik", "hårdrock",
"indierock". 

Example output:
{ "1": ["Ytterplagg för överkroppen, som räcker ungefär till knäet och har ärmar.", "Han hade
på sig en lång rock."], "2": ["Typ av melodisk enkel musik i kraftfullt markerad 4/4-takt.",
"Bandet spelade rock."] }

USER PROMPT:
Entry: öppna
Senses: 1: related to "öppen", "bryta", "bryta upp", "dekantera";
2: related to "starta", "öppnande", "verksamhet"
\end{verbatim}

\subsection{API costs}
\label{app:apicosts}

Table~\ref{table:apicosts} shows the usage cost for some of
the LLMs evaluated in this paper. The table shows the 
cost in US dollars of one API call when the prompt includes
human-written definitions. Neighborhood-based prompts are somewhat
cheaper, and chain-of-thought prompts more expensive.

\begin{table}[t]
\begin{center}
\begin{small}
\begin{tabular}{l c}
\toprule
\textbf{Model} & \textbf{Cost} \\
\midrule
\texttt{gpt-4-turbo} & 0.0053 \\
\texttt{llama-3.1-405b-instruct} (via Replicate) & 0.0036 \\
\texttt{gpt-4o} & 0.0027 \\ 
\texttt{claude-3.5-sonnet} & 0.0026 \\ 
\texttt{llama-3-70b-instruct} (via Replicate) & 0.00037 \\
\texttt{command-r-plus} & 0.0016 \\
\bottomrule
\end{tabular}
\end{small}
\end{center}
\caption{Costs in US dollars per API call for a subset of the models,
with human-written definitions.}
\label{table:apicosts}
\end{table}

\end{document}